\documentclass{article}

    \PassOptionsToPackage{numbers, compress}{natbib}

\usepackage[preprint]{neurips_2023}

\usepackage[utf8]{inputenc} 
\usepackage[T1]{fontenc}    
\usepackage{hyperref}       
\usepackage{url}            
\usepackage{booktabs}       
\usepackage{amsfonts}       
\usepackage{nicefrac}       
\usepackage{microtype}      
\usepackage{xcolor}         
\usepackage{amsmath}
\usepackage{amssymb}
\usepackage{wrapfig}
\usepackage{graphicx}

\title{Multi-CLIP: Contrastive Vision-Language Pre-training for Question Answering tasks in 3D Scenes}

\author{\textbf{Alexandros Delitzas}\footnotemark[1]\textnormal{,}\quad Maria Parelli\footnotemark[1]\textnormal{,}\quad Nikolas Hars\textnormal{,}\quad Georgios Vlassis\textnormal{,}\\\textbf{Sotirios Anagnostidis}\textnormal{,}\quad \textbf{Gregor Bachmann}\textnormal{,}\quad \textbf{Thomas Hofmann}\\
\vspace{-0.35cm}\\
ETH Zurich, Switzerland\\
{\texttt{\{adelitzas, mparelli, nihars, gvlassis, sanagnos, gregorb\}@ethz.ch}}
}

\begin{document}

\maketitle
{
  \renewcommand{\thefootnote}%
    {\fnsymbol{footnote}}
  \footnotetext[1]{These authors contributed equally and are listed alphabetically.}
}

\begin{abstract}
  Training models to apply common-sense linguistic knowledge and visual concepts from 2D images to 3D scene understanding is a promising direction that researchers have only recently started to explore. However, it still remains understudied whether 2D distilled knowledge can provide useful representations for downstream 3D vision-language tasks such as 3D question answering. In this paper, we propose a novel 3D pre-training Vision-Language method, namely Multi-CLIP, that enables a model to learn language-grounded and transferable 3D scene point cloud representations. We leverage the representational power of the CLIP model by maximizing the agreement between the encoded 3D scene features and the corresponding 2D multi-view image and text embeddings in the CLIP space via a contrastive objective. To validate our approach, we consider the challenging downstream tasks of 3D Visual Question Answering (3D-VQA) and 3D Situated Question Answering (3D-SQA). To this end, we develop novel multi-modal transformer-based architectures and we demonstrate how our pre-training method can benefit their performance. Quantitative and qualitative experimental results show that Multi-CLIP outperforms state-of-the-art works across the downstream tasks of 3D-VQA and 3D-SQA and leads to a well-structured 3D scene feature space.
\end{abstract}

\section{Introduction}
\label{sec:introduction}
\begin{wrapfigure}{L}{0.5\textwidth}

\centering
\includegraphics[width=.5\textwidth]{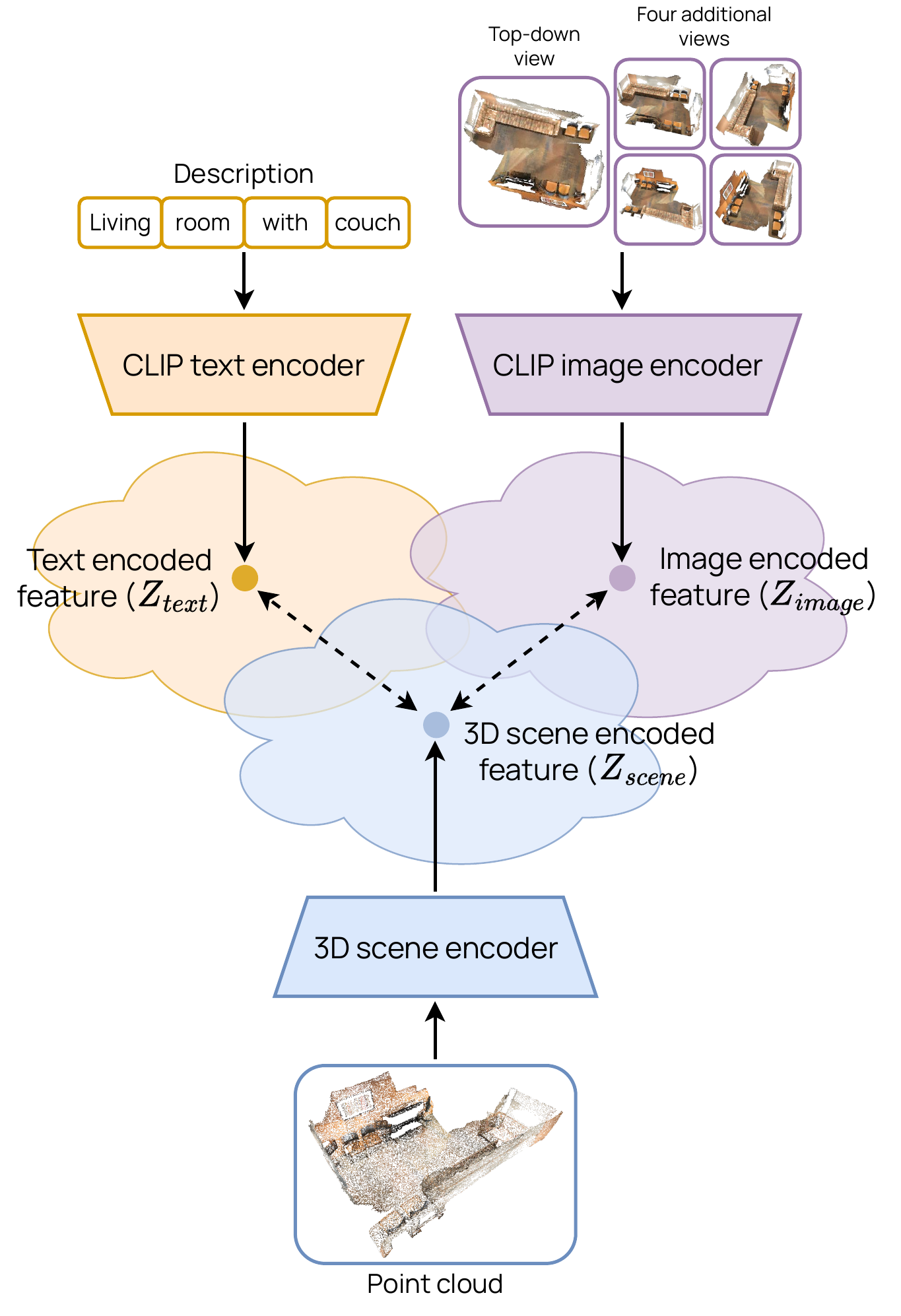}

\caption{Our pre-training method encourages the alignment of the 3D scene representation to the corresponding text and multi-view image embeddings in CLIP space via a contrastive loss.}
\label{fig:pretraining}

\end{wrapfigure}
Humans, by nature, have a coupled representation of textual and visual structures to shape their perception of the world. In recent years, vision and language research has shown remarkable progress in developing models that effectively draw correspondences between the visual and textual modalities. One of the first vision and language tasks, which requires high-level commonsense reasoning and an enhanced textual and visual understanding is Visual Question Answering (VQA)~\cite{antol_2015_vqa,vqacp,matter}, where the model predicts an answer based on the visual content in an image. The dominant approaches in this task leverage elaborate pre-training strategies and multi-modal transformer-based architectures, which learn generalizable contextualized text and visual embeddings. However, these methods are restricted to 2D visual information and this limitation affects their usability in many 3D world applications, such as robotics and AR/VR.

In this direction, recent works aim to extend VQA to the 3D domain. Two recently released benchmarks, namely ScanQA~\cite{azuma_2022_scanqa} and SQA3D~\cite{ma_2022_sqa3d}, introduce the tasks of 3D Visual Question Answering (3D-VQA) and 3D Situated Question Answering (3D-SQA) respectively. In the 3D-VQA setting, a model receives visual information of a 3D scene from a rich RGB-D indoor scan and has to answer a given question about the scene content and localize the referred objects. In the 3D-SQA setting, an additional textual input is provided describing the situation (i.e., position and orientation) of the agent in the scene. The task is to first understand the agent's situation and then answer a question about the surrounding environment. However, extending 2D VQA methods to the 3D domain is not trivial, since 3D models face the challenge of understanding complex geometry structures and spatial relations among objects. 

To endow models with 3D semantic understanding and reasoning abilities, a new line of research has emerged. Recent works~\cite{Peng2023OpenScene, zhang2023clip, PointCLIP} harness 2D knowledge of foundation models, such as CLIP~\cite{radford_2021_clip}, and achieve state-of-the-art zero-shot performance on the tasks of 3D object recognition and 3D semantic segmentation. Other lines of work on 3D representation learning tackle these tasks by leveraging pre-training strategies. In \cite{liu_2021_learnfrom2d,Afham_2022_CVPR}, a contrastive pixel-to-point approach is employed to transfer 2D visual information to 3D models, while in \cite{zhang_2021_depth_contrast}, an unsupervised pre-training scheme is explored to align 3D point clouds and voxel representations. However, to our knowledge, no pretext methods for vision-language tasks, such as 3D question answering, have been proposed that guide a model to correlate scene-level 3D visual input to language cues and 2D information.

In this work, we propose Multi-CLIP, a simple yet effective 3D Vision-Language (V-L) pre-training method to help a model learn transferable and semantically meaningful scene representations. We demonstrate that infusing knowledge from large-scale pre-trained 2D vision and language models can enhance a model’s 3D scene-based perception and boost performance in downstream 3D vision-language tasks. To this end, we develop a transformer-based 3D scene encoder module, which captures a holistic 3D scene representation by modeling the spatial relations between the scene's object features. Our proposed pre-training objective is to facilitate 3D point cloud understanding by training the scene encoder to project the appearance and geometric features of the 3D scan to an interpretable embedding space with desirable properties. This can be achieved by aligning the 3D scene-aware embeddings to the corresponding text and multi-view image representations generated by CLIP~\cite{radford_2021_clip}.

To validate the effectiveness of our approach, we consider the challenging downstream tasks of 3D-VQA and 3D-SQA. To this end, we transfer the weights of the pre-trained 3D scene encoder to a novel 3D Vision-Linguistic architecture that fuses the multi-modal representations and fine-tune the model in a supervised manner. Our fine-tuned model achieves state-of-the-art results on the tasks of 3D-VQA and 3D-SQA as well as on the auxiliary task of referred object localization. We also provide a visualization of the learned 3D scene features after pre-training, demonstrating that our model learns mappings in a semantically-structured latent space, where scenes with similar properties are clustered together.

\section{Related Work}

\textbf{3D Question Answering.} The task that initiated research on language-driven 3D scene understanding was 3D language grounding, in which a model has to localize an object in a scene guided by a textual description \cite{chen_2020_scanrefer, thomason_2021_3d_language_grounding, achlioptas2020referit_3d,cai20223djcg}. Building upon these advances, a novel task has been proposed, namely 3D Visual Question Answering, which tests a model’s spatial relation comprehension and commonsense reasoning abilities. In this problem, a model receives 3D visual information, often in the form of a 3D scene scan, and is tasked with answering a question about the scene. A few benchmarks have been proposed in this direction, such as \cite{azuma_2022_scanqa}, which introduces a new 3D-VQA dataset, namely ScanQA, based on ScanNet~\cite{dai_2017_scannet} scenes and develops an architecture that jointly models 3D object and question features to predict the correct answer. More recently, Ma et al.~\cite{ma_2022_sqa3d} proposed SQA3D, a dataset for embodied scene understanding and question answering. In this setting, the agent has to localize its situation in the 3D scene as described by a textual prompt and answer a question about its environment. In this work, we develop novel architectures to tackle the tasks of 3D-VQA and 3D-SQA and show that our pre-training method leads to state-of-the-art results on the ScanQA and SQA3D benchmarks.

\textbf{2D and 3D Vision-Language Pre-training.} In the 2D domain, V-L pre-training has been thoroughly investigated \cite{chen_2020_uniter,li_2020_oscar,Su2020VLBERT,huang2021seeing,yang2022vision}. Methods for language and image comprehension tasks have largely benefited from extensive pre-training on large-scale V-L datasets, enabling meaningful image-text representations to be extracted. Two characteristic examples are VL-BERT~\cite{Su2020VLBERT} and UNITER~\cite{chen_2020_uniter}. The core of these architectures is a multimodal transformer encoder, which leverages enriched pre-training strategies for global image-text alignment and fine-grained word-region mapping.

In the 3D domain, current state-of-the-art pre-training methods have focused on learning enhanced 3D point cloud semantic attributes that can be successfully deployed in downstream tasks, such as 3D object classification and scene segmentation~\cite{PointContrast2020,rozenberszki2022language,zhang_2021_depth_contrast}. Initial lines of work focused on learning transferable and augmentation-invariant point cloud representations.  In DepthContrast~\cite{zhang_2021_depth_contrast}, the authors propose a joint pre-training framework of point cloud and voxel architectures via an instance discrimination objective. More recent lines of work employ cross-modal pre-training methods, by exploiting large-scale 2D models. In \cite{liu_2021_learnfrom2d}, the authors map pixel-level and point-level features into the same latent space using a pixel-to-point contrastive loss. Similarly, in CrossPoint~\cite{Afham_2022_CVPR}, the authors enforce 3D-2D correspondences of object features and invariance to affine point cloud transformations via self-supervised contrastive learning. 
However, these approaches investigate pre-training in the context of 2D images or focus on single objects and lack scene context. To our knowledge, pre-training methods for language-based 3D scene reasoning and question-answering tasks, which leverage the synergies between textual, 2D and 3D visual modalities have not been explored and with this work we aim to stimulate interest in this direction.

\section{Proposed Method}
\label{sec:proposedmethod}
We propose a pretext method (Figure~\ref{fig:pretraining}) that aligns the 3D scene representation to the corresponding text and multi-view image embeddings in CLIP space~\cite{radford_2021_clip} via a contrastive learning loss. To demonstrate its effectiveness, we (a) pre-train a 3D scene encoder with this objective (Section~\ref{sec:pre-training-method}) and (b) transfer the learned representations to a novel 3D V-L model and fine-tune it for the downstream tasks of 3D-VQA (Section~\ref{sec:3d-vqa-method}) and 3D-SQA (Section~\ref{sec:3d-sqa-method}). Since the problem setting of the two downstream tasks differs, we slightly modify the architecture of the 3D V-L model to train for each task independently.

\subsection{Pre-training Framework Overview}
\label{sec:pre-training-method}

\begin{figure}
\centering
\includegraphics[width=0.99\textwidth]{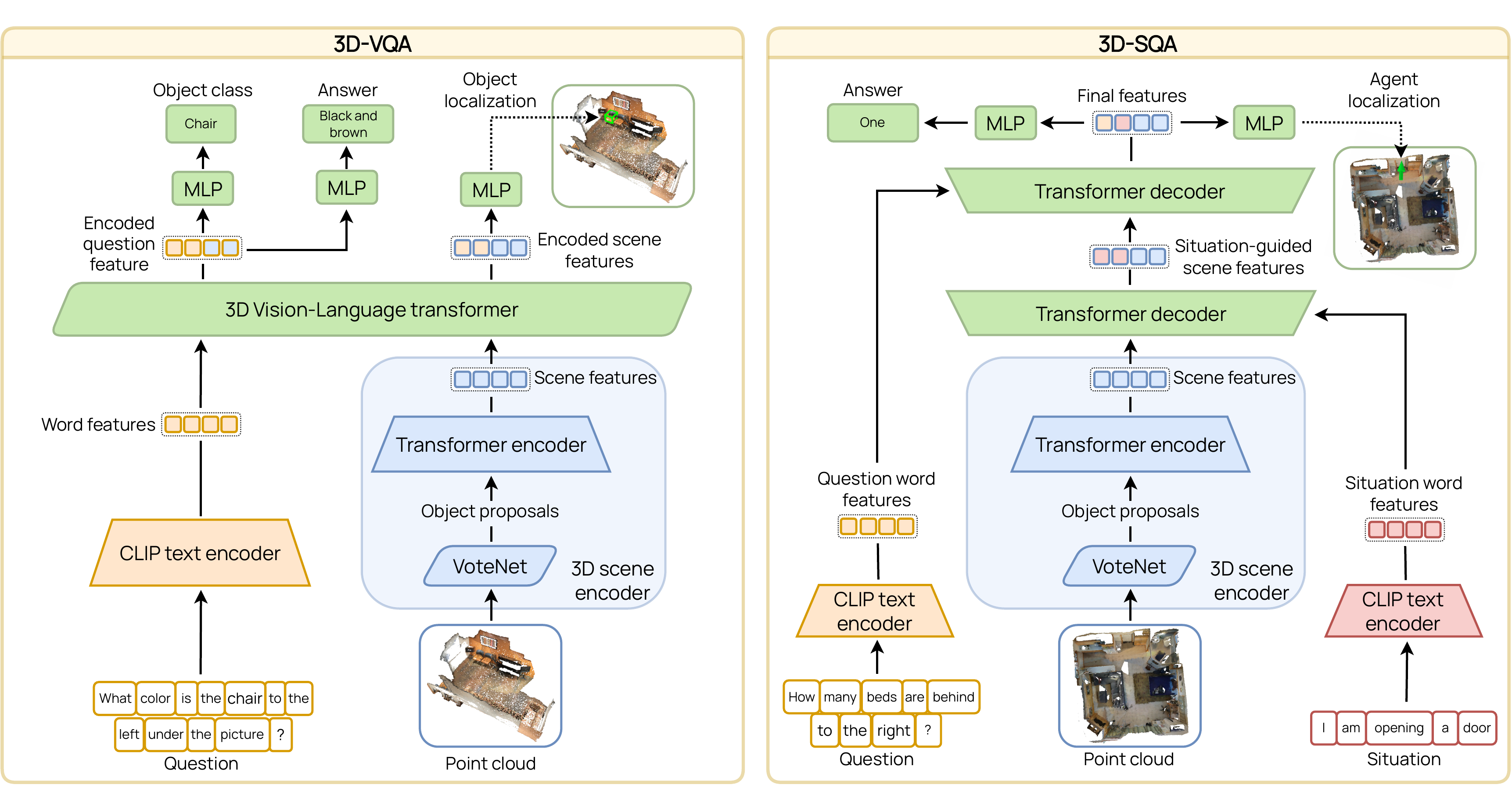}
\caption{The proposed model architectures for the two downstream tasks of 3D-VQA (left) and 3D-SQA (right).}
\label{fig:architecture}
\end{figure}

\textbf{3D Scene Encoder.}
We model the scene as an RGB-colored point cloud of $N$ points, $p \in \mathbb{R}^{N\times 6}$, which is processed by the 3D scene encoder to generate a holistic representation of the scene objects. The scene encoder $f_{\theta}$ comprises a VoteNet~\cite{qi_2019_votenet} with a pre-trained PointNet++~\cite{zhang_2021_depth_contrast} backbone, which outputs a set of $M$ point clusters, $\mathbf{C}_v\in \mathbb{R}^{M\times 128}$, representing 3D object proposals with enriched point features. The VoteNet model is followed by a transformer encoder layer~\cite{Vaswani2017}, which leverages self-attention to refine the object representations by modeling their spatial relations. We denote the final scene representation obtained by our scene encoder $f_{\theta}$ as $\mathbf{C} \in \mathbb{R}^{M\times 128}$.


\textbf{3D Scene Encoder Pre-training.}
The examined tasks of 3D-VQA and 3D-SQA require commonsense reasoning and a solid understanding of the underlying relations among the text and 3D object regions in the scene. 
Thus, the objective of our pre-training method is to encourage discriminative feature learning by transferring the rich 2D visual and linguistic understanding of CLIP to the 3D model. 
To achieve this, we aim to maximize the agreement between the scene-level features, generated by our 3D scene encoder $f_{\theta}$, to the corresponding  CLIP text and multi-view image representations.
In order to learn an enhanced 2D representation of the scene, we render it from five different views through a rotation along the z-axis with equal angles. These views are processed by the CLIP image encoder and we fuse the multi-view features by averaging them to obtain a view-robust scene representation $\mathbf{Z}_{image}$. We embed the textual description of the scene to the same feature space via CLIP’s text encoder, receiving the text representation $\mathbf{Z}_{text}$.

To generate the 3D scene global representation $\mathbf{Z}_{scene}$, we follow the practice of \cite{dosovitskiy2020vit} and append a learnable classification token to the input sequence of object features extracted by VoteNet. The state of this token (i.e., global feature) at the output of the transformer encoder is projected to the CLIP feature space in $\mathbb{R}^d$ via a linear projection head.
Our goal is to align the text, image and point cloud modalities. To achieve this, we leverage a 3D-2D and 3D-text Noise-Contrastive Estimation (NCE) loss for contrastive learning by modifying the InfoNCE loss~\cite{oord2019representation}. Our proposed loss encourages the model to bring 3D scene features close to their corresponding 2D image and text features while separating them from other dissimilar 2D image and text embeddings. Formally,

\begin{equation}
\mathcal{L}_{image} = -\frac{1}{|B|} \sum_{i=1}^B \log \frac{\exp \left(\mathbf{Z}_{i,scene}^{\top} \mathbf{Z}_{i,image} / \tau\right)}{\sum_{j = 1}^B \exp \left(\mathbf{Z}_{i,scene}^{\top} \mathbf{Z}_{j,image} / \tau\right)}
\end{equation}
\begin{equation}
\mathcal{L}_{text} = -\frac{1}{|B|} \sum_{i=1}^B \log \frac{\exp \left(\mathbf{Z}_{i,scene}^{\top} \mathbf{Z}_{i,text} / \tau\right)}{\sum_{j = 1}^B \exp \left(\mathbf{Z}_{i,scene}^{\top} \mathbf{Z}_{j,text} / \tau\right)}
\end{equation}
where $(\mathbf{Z}_{i,scene}, \mathbf{Z}_{i,text})$ and $(\mathbf{Z}_{i,scene}, \mathbf{Z}_{i,image})$ are the positive text-scene and image-scene pairs for each sample $i$ of a batch respectively, $B$ is the batch size and $\tau$ is the temperature coefficient.

Following previous works~\cite{azuma_2022_scanqa, ma_2022_sqa3d}, we add an auxiliary loss term that supervises the VoteNet module to facilitate 3D object detection and classification, which we denote as $\mathcal{L}_{det}$. This loss term helps the VoteNet module of our 3D scene encoder to produce accurate object proposals, which contribute to a refined global scene representation. We refer the reader to the supplementary material for more details about the loss term $\mathcal{L}_{det}$.

Formally, the final loss for the pre-training is defined as 

\begin{equation}
    \mathcal{L}_{pre} = \mathcal{L}_{det} + \alpha \mathcal{L}_{text} + \beta \mathcal{L}_{image}
\end{equation}
where we set $ \alpha, \beta = 0.5$.

\subsection{Model Architecture for 3D-VQA}
\label{sec:3d-vqa-method}
Our architecture for the downstream 3D-VQA task (Figure~\ref{fig:architecture}) consists of three modules, the pre-trained 3D scene encoder $f_{\theta}$, which processes the 3D scene point features, a CLIP text encoder that generates the question word embeddings and a 3D Vision-Language Transformer that fuses the visual and question representations. The model is tasked with finding the correct answer to the question and localizing the target object referred to by the question.

To process the question, we use CLIP’s text encoder to obtain $512$-dimensional word-level embeddings $\mathbf{Q} \in \mathbb{R}^{N_q\times 512}$, where $N_q$ is the number of words in the question. The word and 3D scene features, extracted by the scene encoder $f_{\theta}$, are transformed to the same hidden dimension $h$ using two independent linear layers. Then, they are concatenated and fused via a two-layer transformer encoder that leverages self-attention to model intra- and inter-modal relations. The updated scene object features $\mathbf{C'} \in \mathbb{R}^{M \times h}$ are forwarded to a linear layer that performs target object localization by determining the likelihood of each of the $M$ object boxes being related to the question. Following CLIP, we treat the updated EOT embedding (last token in the sequence) of the question as the pooled question feature $\mathbf{Q'} \in \mathbb{R}^{h}$ and use it as input to two linear classifiers. The first one predicts the correct answer by projecting $\mathbf{Q'}$ into a vector $a \in \mathbb{R}^{N_a}$  for the $N_a$ answer candidates. The second predicts which objects from the 18 ScanNet~\cite{dai_2017_scannet} classes are associated with the question.

\textbf{Loss Function.}
We model the final loss as a linear combination of four terms. We utilize the referred object localization loss $\mathcal{L}_{loc}$, as defined in \cite{chen_2020_scanrefer}, and the object detection loss $\mathcal{L}_{det}$  of VoteNet \cite{qi_2019_votenet}. To further supervise the training, we include an object classification loss $\mathcal{L}_{obj}$, which is modeled as a multi-class cross-entropy loss and an answer classification loss $\mathcal{L}_{ans}$, which is a binary cross-entropy (BCE) loss function as there may be multiple candidate answers.
Therefore, the total loss is defined as $\mathcal{L}_{vqa} = \mathcal{L}_{det} + \mathcal{L}_{obj} + \mathcal{L}_{ans} + \mathcal{L}_{loc}$.

\subsection{Model Architecture for 3D-SQA}
\label{sec:3d-sqa-method}
In 3D-SQA, the model is given an additional input sentence that describes the situation of an agent in the scene and answers a relevant question. Thus, we modify our architecture (Figure~\ref{fig:architecture}) to incorporate the situation into our pipeline. We leverage CLIP’s text encoder to obtain $512$-dimensional word-level situation embeddings $\mathbf{S} \in \mathbb{R}^{N_s\times 512}$, where $N_s$ is the number of words in the situation description. The situation word embeddings $\mathbf{S}$ are used as input to a transformer decoder (functioning as query tokens) and the 3D scene features $\mathbf{C}$ are used to generate the keys and values. We take advantage of the cross-attention module to capture the relationships between them and generate object-centric feature tokens guided by the situation. These are forwarded to a second transformer decoder, where they attend to the language tokens of the question $\mathbf{Q}$ to yield the final representation. Finally, we utilize two MLPs, one for predicting the answer $a \in \mathbb{R}^{N_a}$ to the question and one for predicting the location $l=(s^{pos}, s^{rot})$ of the agent in the scene.

\textbf{Loss Function.}
We adopt a similar loss formulation as the one described in the previous section. In order to adapt it to the 3D-SQA task, we substitute the object localization and classification loss with two auxiliary MSE losses $\mathcal{L}_{pos}$ and $\mathcal{L}_{rot}$ that encourage the model to accurately predict the position $s^{pos}$ and orientation $s^{rot}$ of the agent in the described input situation respectively. The orientation is represented as quaternion $(x, y, z, w)$ and the position as a 3D coordinate $(x, y, z)$. 
Thus, the total loss is defined as $\mathcal{L}_{sqa} = \mathcal{L}_{det} + \mathcal{L}_{ans} + \mathcal{L}_{pos} + \mathcal{L}_{rot}$.

\section{Experiments}
\label{sec:experiments}

In this section, we validate our method by transferring the learned representations of our pre-trained 3D network to two downstream 3D visual-linguistic tasks. The tasks for evaluation are (a) 3D visual question answering and the auxiliary task of referred object localization on the ScanQA dataset and (2) 3D situated question  answering on the SQA3D dataset.
\setlength{\tabcolsep}{3pt}
\setlength{\abovetopsep}{1ex}
\begin{table*}
  \caption{Comparison of 3D visual question answering results on the ScanQA test datasets.}
  \label{tab:results_vqa_acc}
  \centering
  \begin{tabular}{@{}lcccccc@{}}
    \toprule
    Method & EM@1 & BLEU-1 & BLEU-4 & ROUGE & METEOR & CIDEr\\
    \midrule
    \textbf{Test set w/ objects} \\
    Scanrefer + MCAN & 20.56 & 27.85 & 7.46 & 30.68 & 11.97 & 57.36\\
    ScanQA & 23.45 & 31.56 & 12.04 & 34.34 & 13.55 & 67.29 \\
    Ours w/o pre-training & 22.76 & 31.08 & \textbf{13.31} & 33.84& 13.28 & 65.81 \\
    Ours & \textbf{24.02} & \textbf{32.63} & 12.65  & \textbf{35.46} & \textbf{13.97} & \textbf{68.70}\\
    \midrule
    \textbf{Test set w/o objects} \\
    Scanrefer + MCAN & 19.04 & 26.98 & 7.82 & 28.61 & 11.38 & 53.41\\
    ScanQA & 20.90 & 30.68 & 10.75 & 31.09 & 12.59 & 60.24 \\
    Ours w/o pre-training & 20.71 & 31.22 & 11.49 & 31.35& 12.80 & 60.75 \\
    Ours & \textbf{21.48} & \textbf{32.69} & \textbf{12.87}  & \textbf{32.61} & \textbf{13.36} & \textbf{63.20}\\
    \bottomrule
  \end{tabular}
\end{table*}

\subsection{Experimental setup}
\textbf{Datasets.}
The ScanQA~\cite{azuma_2022_scanqa} dataset contains 41,363 diverse question-answer pairs and 3D object localization annotations for 800 indoor 3D scenes of the ScanNet~\cite{dai_2017_scannet} dataset. ScanQA also includes two test sets with and without object annotations. The SQA3D~\cite{ma_2022_sqa3d} dataset provides around 6,800 unique situations, based upon 650 ScanNet scenes, accompanied by 20,400 descriptions and 33,400 diverse reasoning questions for these situations. ScanNet~\cite{dai_2017_scannet} is a large-scale annotated dataset of 3D mesh reconstructions of interior spaces. In the pre-training phase, we obtain the textual descriptions using the ScanRefer dataset~\cite{chen_2020_scanrefer}, which provides 51,583 descriptions of 800 ScanNet scenes. We also use the ScanNet point cloud data to render the RGB images from multiple views with the Open3D software~\cite{zhou_2018_open3d}.

\setlength{\tabcolsep}{3pt}
\setlength{\abovetopsep}{1ex}
\begin{wraptable}{R}{0.5\textwidth}
\caption{Comparison of referred object localization results on the ScanQA valid dataset.}
  \label{tab:results_loc}
  \centering
  \begin{tabular}{@{}lcc@{}}
    \toprule
    Method & Acc@0.25 & Acc@0.5 \\
    \midrule
    Scanrefer + MCAN & 23.53 & 11.76\\
    ScanQA & 24.96 & 15.42 \\
    Ours w/o pre-training & 26.57 & 18.58 \\
    Ours & \textbf{29.60} & \textbf{21.41} \\
    \bottomrule
  \end{tabular}
\end{wraptable}

\textbf{Implementation Details.}
We pre-train the 3D scene encoder for 15K iterations with Adam~\cite{kingma2014adam} optimizer using a batch size of 16, a learning rate of 1e-4 and a weight decay of 1e-5. We use the pre-trained weights of the scene encoder and fine-tune the 3D-VQA network on ScanQA for 40 epochs with an initial learning rate of 5e-4, which we decrease by a factor of 0.2 in epoch 15. Likewise, we fine-tune the 3D-SQA network on SQA3D for 50 epochs with the same initial learning rate. To mitigate overfitting, we applied rotation about all three axes using a random angle in $\left[-5^{\circ}, 5^{\circ}\right]$ and randomly translated the point cloud within 0.5 m in all directions. We also used random cuboid augmentation, similar to \cite{3detr}, which extracts random cuboids from the input point cloud.

\subsection{Results}
\label{sec:results}
\begin{figure}
\centering
\includegraphics[width=0.9\textwidth]{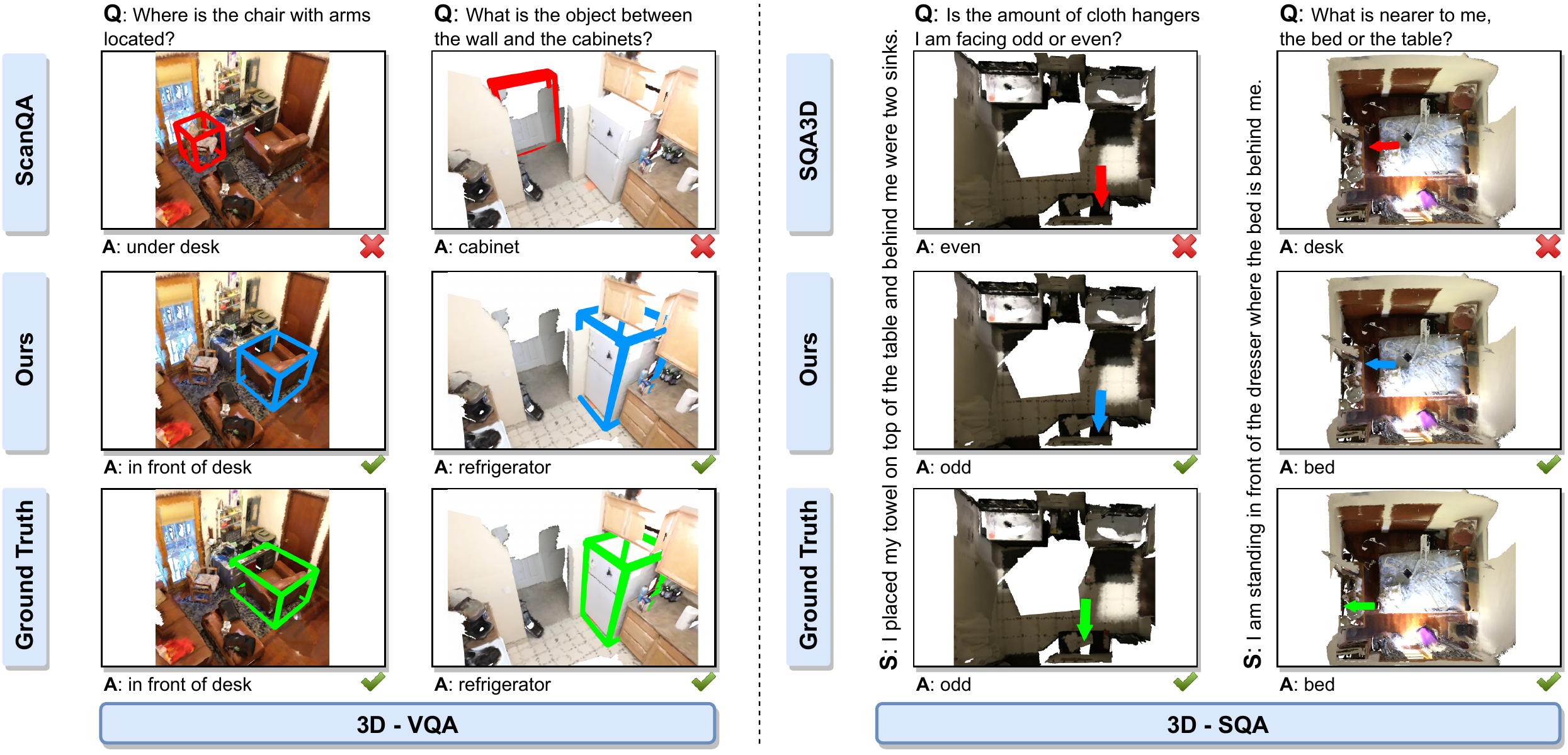}
\caption{Qualitative results on the ScanQA (left) and SQA3D dataset (right).}
\label{fig:qualitative_results}
\end{figure}

\setlength{\tabcolsep}{3pt}
\setlength{\abovetopsep}{1ex}
\begin{wraptable}{R}{0.45\textwidth}
\caption{Comparison of 3D situated question answering results on the SQA3D test dataset.}
  \label{tab:results_sqa_acc}
  \centering
  \begin{tabular}{@{}lc@{}}
    \toprule
    Method & EM@1 \\
    \midrule
    SQA3D & 47.20 \\
    Ours w/o pre-training & 47.38 \\
    Ours & \textbf{48.02}  \\
    \bottomrule
  \end{tabular}  
\end{wraptable}
\textbf{3D Visual Question Answering.}
To measure the downstream performance of our model on 3D-VQA, we report the EM@1 metric, which is the percentage of predictions in which the predicted answer exactly matches any of the ground-truth answers. Following the practice of \cite{azuma_2022_scanqa}, we include the sentence evaluation metrics BLEU~\cite{bleu}, ROUGE-L~\cite{rouge}, METEOR~\cite{meteor} and CIDEr~\cite{cider}. These metrics are significant for evaluating robust answer matching since some questions have multiple possible answer expressions. To assess the referred object localization accuracy, we report the Acc@0.25 and Acc@0.5 metrics, which are the percentage of bounding box predictions that have a higher IoU with the ground truths than the threshold 0.25 and 0.5 respectively. 
As baselines, we use the current state-of-the-art method of ScanQA~\cite{azuma_2022_scanqa} as well as ScanRefer+MCAN~\cite{mcan}, where a pre-trained ScanRefer~\cite{chen_2020_scanrefer} model identifies the referred object and the MCAN model is applied to the image surrounding it. We also compare to the performance of our model trained from scratch. The results are displayed in Table~\ref{tab:results_vqa_acc} and Table~\ref{tab:results_loc}. With our pre-training method, we report a significant increase in the question answering metrics and a gain of $3.03\%$ and $2.83\%$ in the Acc@0.25 and Acc@0.5 metrics, respectively. This validates the effectiveness of our pre-training strategy in both question answering and referred object localization. We also observe that our method achieves a notable improvement over the ScanQA baseline.

\textbf{3D Situated Question Answering.}
We further evaluate our method on the SQA3D test set. Following the practice of \cite{ma_2022_sqa3d}, we adopt the EM@1 as our evaluation metric. We report our results in Table~\ref{tab:results_sqa_acc}, where we use the current state-of-the-art method of \cite{ma_2022_sqa3d} as a baseline. After pre-training and fine-tuning on the SQA3D train split, we report clear performance gains in answer accuracy compared to training from scratch.

\subsection{Visualization}

\begin{figure}
\centering
\includegraphics[width=0.9\textwidth]{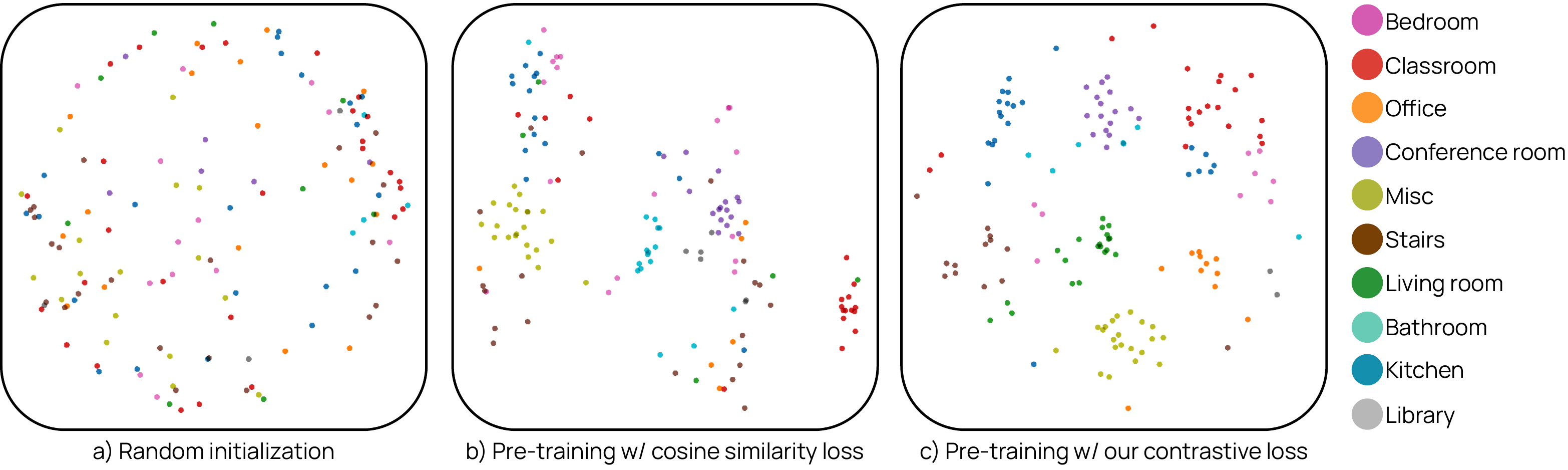}
\caption{T-SNE visualizations of scene-level features in ScanNet. The 3D scene encoder weights learned during pre-training lead to a structured feature representation space.}
\label{fig:t-sne}
\end{figure}

In Figure~\ref{fig:t-sne}, we provide the T-SNE~\cite{maaten_2008_t-sne} visualization of the learned features of the pre-trained 3D scene encoder without fine-tuning on downstream tasks. We observe that scenes with semantically similar properties (i.e., same scene type) form clusters in the embedding space. This highlights the high-level semantic understanding ability acquired by the model when it is pre-trained with rich 2D visual and linguistic information. Additionally, we observe that our proposed contrastive pre-training objective leads to more discriminative representations in feature space than using a commonly used cosine similarity loss to align 3D-2D and 3D-text features, as later discussed in Section~\ref{sec:ablation}.

\subsection{Ablation study}
\label{sec:ablation}

\setlength{\tabcolsep}{3pt}
\setlength{\abovetopsep}{1ex}
\begin{wraptable}{R}{0.4\textwidth}
    \caption{Ablation study on the SQA3D test set.}
    \label{tab:ablation}
  \centering
  \begin{tabular}{@{}lc@{}}
    \toprule
    Method & EM@1  \\
    \midrule
    Ours w/o multi-view & 47.81 \\
    Ours w/ cosine similarity & 47.73  \\
    Ours w/o $\mathcal{L}_{text}$ & 47.59  \\
    Ours w/o $\mathcal{L}_{image}$ & 47.65  \\
    Ours (full) & \textbf{48.02}  \\
    \bottomrule
  \end{tabular}
\end{wraptable}

To justify our pre-training pipeline design choices, we include ablation studies (Table~\ref{tab:ablation}) on SQA3D.

\textbf{Do multiple views help?} Many works have confirmed that learning multi-modal representations that are robust to view changes benefits 3D object and scene understanding. Thus, we hypothesize that rendering the 3D scene point cloud to several 2D planes and aggregating the multiview information will induce performance gains in the examined downstream tasks. We compare the result of pre-training with 5 views to training only with the top-down view of the scene. It can be observed that adding more views increases accuracy. This confirms our intuition that a multi-view representation can facilitate correspondence learning between 3D scene and language queries. 

\textbf{Is contrastive loss a good choice?} We investigate experimentally whether contrastive learning is the optimal choice for our pre-training strategy in this setting by substituting the contrastive objective with a cosine similarity distance loss. As we can infer from both experimental results in Table~\ref{tab:ablation} and feature space visualizations in Figure~\ref{fig:t-sne}, contrastive learning leads to increased downstream performance and encourages the model to capture representations that discriminate more effectively between similar and dissimilar data points.

\textbf{Do all loss terms matter?} We explore the added benefit of each term in our contrastive loss formulation. The results show that removing either $\mathcal{L}_{text}$ or $\mathcal{L}_{image}$ degrades accuracy. This validates our hypothesis that connecting 3D point clouds to both their language descriptions and 2D multi-view images promotes spatial relations comprehension and reasoning.

\section{Conclusion}
\label{sec:conclusion}

In this paper, we propose Multi-CLIP, a novel V-L pre-training strategy that helps a model learn semantically meaningful and language-grounded 3D scene features which can be transferred to 3D scene reasoning and question answering downstream tasks. This is achieved by aligning the 3D extracted features to the corresponding captions and rendered 2D multi-view images in the CLIP embedding space via a contrastive objective. Our quantitative and qualitative results on the downstream tasks of 3D-VQA and 3D-SQA show the effectiveness of our method in learning rich 3D scene representations and demonstrate state-of-the-art performance on the challenging ScanQA and SQA3D benchmarks.

{
\small
\bibliographystyle{unsrt}
\bibliography{egbib}

\begin{thebibliography}{10}

\bibitem{antol_2015_vqa}
Stanislaw Antol, Aishwarya Agrawal, Jiasen Lu, Margaret Mitchell, Dhruv Batra,
  C.~Lawrence Zitnick, and Devi Parikh.
\newblock {VQA}: {V}isual {Q}uestion {A}nswering.
\newblock In {\em International Conference on Computer Vision (ICCV)}, 2015.

\bibitem{vqacp}
Aishwarya Agrawal, Dhruv Batra, Devi Parikh, and Aniruddha Kembhavi.
\newblock Don't just assume; look and answer: Overcoming priors for visual
  question answering.
\newblock In {\em Proceedings of the IEEE Conference on Computer Vision and
  Pattern Recognition (CVPR)}, 2018.

\bibitem{matter}
Yash Goyal, Tejas Khot, Douglas Summers-Stay, Dhruv Batra, and Devi Parikh.
\newblock Making the v in vqa matter: Elevating the role of image understanding
  in visual question answering.
\newblock {\em International Journal of Computer Vision}, 127:398--414, 2016.

\bibitem{azuma_2022_scanqa}
Daichi Azuma, Taiki Miyanishi, Shuhei Kurita, and Motoki Kawanabe.
\newblock {ScanQA: 3D} question answering for spatial scene understanding.
\newblock In {\em {Proceedings of the IEEE/CVF Conference on Computer Vision
  and Pattern Recognition (CVPR)}}, 2022.

\bibitem{ma_2022_sqa3d}
Xiaojian Ma, Silong Yong, Zilong Zheng, Qing Li, Yitao Liang, Song-Chun Zhu,
  and Siyuan Huang.
\newblock Sqa3d: Situated question answering in 3d scenes.
\newblock In {\em International Conference on Learning Representations}, 2023.

\bibitem{Peng2023OpenScene}
Songyou Peng, Kyle Genova, Chiyu~"Max" Jiang, Andrea Tagliasacchi, Marc
  Pollefeys, and Thomas Funkhouser.
\newblock Openscene: 3d scene understanding with open vocabularies.
\newblock In {\em Proceedings of the IEEE/CVF Conference on Computer Vision and
  Pattern Recognition (CVPR)}, 2023.

\bibitem{zhang2023clip}
Junbo Zhang, Runpei Dong, and Kaisheng Ma.
\newblock Clip-fo3d: Learning free open-world 3d scene representations from 2d
  dense clip.
\newblock {\em arXiv preprint arXiv:2303.04748}, 2023.

\bibitem{PointCLIP}
Renrui Zhang, Ziyu Guo, Wei Zhang, Kunchang Li, Xupeng Miao, Bin Cui, Yu~Qiao,
  Peng Gao, and Hongsheng Li.
\newblock Pointclip: Point cloud understanding by clip.
\newblock In {\em 2022 IEEE/CVF Conference on Computer Vision and Pattern
  Recognition (CVPR)}, pages 8542--8552, 2022.

\bibitem{radford_2021_clip}
Alec Radford, Jong~Wook Kim, Chris Hallacy, Aditya Ramesh, Gabriel Goh,
  Sandhini Agarwal, Girish Sastry, Amanda Askell, Pamela Mishkin, Jack Clark,
  Gretchen Krueger, and Ilya Sutskever.
\newblock {Learning} transferable visual models from natural language
  supervision.
\newblock {\em CoRR}, abs/2103.00020, 2021.

\bibitem{liu_2021_learnfrom2d}
Yueh{-}Cheng Liu, Yu{-}Kai Huang, HungYueh Chiang, Hung{-}Ting Su, Zhe~Yu Liu,
  Chin{-}Tang Chen, Ching{-}Yu Tseng, and Winston~H. Hsu.
\newblock {Learning from 2D: Pixel}-to-point knowledge transfer for {3D}
  pretraining.
\newblock {\em CoRR}, abs/2104.04687, 2021.

\bibitem{Afham_2022_CVPR}
Mohamed Afham, Isuru Dissanayake, Dinithi Dissanayake, Amaya Dharmasiri,
  Kanchana Thilakarathna, and Ranga Rodrigo.
\newblock Crosspoint: Self-supervised cross-modal contrastive learning for 3d
  point cloud understanding.
\newblock In {\em Proceedings of the IEEE/CVF Conference on Computer Vision and
  Pattern Recognition (CVPR)}, pages 9902--9912, June 2022.

\bibitem{zhang_2021_depth_contrast}
Zaiwei Zhang, Rohit Girdhar, Armand Joulin, and Ishan Misra.
\newblock Self-supervised pretraining of 3d features on any point-cloud.
\newblock In {\em 2021 IEEE/CVF International Conference on Computer Vision
  (ICCV)}, pages 10232--10243, 2021.

\bibitem{chen_2020_scanrefer}
Dave~Zhenyu Chen, Angel~X Chang, and Matthias Nie{\ss}ner.
\newblock {Scanrefer: 3D} object localization in {RGB-D} scans using natural
  language.
\newblock In {\em {Computer Vision--ECCV 2020: 16th European Conference,
  Glasgow, UK, August 23--28, 2020, Proceedings, Part XX 16}}, pages 202--221.
  Springer, 2020.

\bibitem{thomason_2021_3d_language_grounding}
Jesse Thomason, Mohit Shridhar, Yonatan Bisk, Chris Paxton, and Luke
  Zettlemoyer.
\newblock {Language} grounding with {3D} objects.
\newblock {\em CoRR}, abs/2107.12514, 2021.

\bibitem{achlioptas2020referit_3d}
Panos Achlioptas, Ahmed Abdelreheem, Fei Xia, Mohamed Elhoseiny, and
  Leonidas~J. Guibas.
\newblock {ReferIt3D}: Neural listeners for fine-grained 3d object
  identification in real-world scenes.
\newblock In {\em 16th European Conference on Computer Vision (ECCV)}, 2020.

\bibitem{cai20223djcg}
Daigang Cai, Lichen Zhao, Jing Zhang, Lu~Sheng, and Dong Xu.
\newblock 3djcg: A unified framework for joint dense captioning and visual
  grounding on 3d point clouds.
\newblock In {\em Proceedings of the IEEE/CVF Conference on Computer Vision and
  Pattern Recognition}, pages 16464--16473, 2022.

\bibitem{dai_2017_scannet}
Angela Dai, Angel~X Chang, Manolis Savva, Maciej Halber, Thomas Funkhouser, and
  Matthias Nie{\ss}ner.
\newblock {Scannet: Richly-annotated 3D} reconstructions of indoor scenes.
\newblock In {\em {Proceedings of the IEEE Conference on Computer Vision and
  Pattern Recognition}}, pages 5828--5839, 2017.

\bibitem{chen_2020_uniter}
Yen-Chun Chen, Linjie Li, Licheng Yu, Ahmed~El Kholy, Faisal Ahmed, Zhe Gan,
  Yu~Cheng, and Jingjing Liu.
\newblock {Uniter: Universal} image-text representation learning.
\newblock In {\em ECCV}, 2020.

\bibitem{li_2020_oscar}
Xiujun Li, Xi~Yin, Chunyuan Li, Xiaowei Hu, Pengchuan Zhang, Lei Zhang, Lijuan
  Wang, Houdong Hu, Li~Dong, Furu Wei, Yejin Choi, and Jianfeng Gao.
\newblock {Oscar: Object}-semantics aligned pre-training for vision-language
  tasks.
\newblock {\em ECCV 2020}, 2020.

\bibitem{Su2020VLBERT}
Weijie Su, Xizhou Zhu, Yue Cao, Bin Li, Lewei Lu, Furu Wei, and Jifeng Dai.
\newblock Vl-bert: Pre-training of generic visual-linguistic representations.
\newblock In {\em International Conference on Learning Representations}, 2020.

\bibitem{huang2021seeing}
Zhicheng Huang, Zhaoyang Zeng, Yupan Huang, Bei Liu, Dongmei Fu, and Jianlong
  Fu.
\newblock Seeing out of the box: End-to-end pre-training for vision-language
  representation learning.
\newblock In {\em The IEEE Conference on Computer Vision and Pattern
  Recognition (CVPR)}, 2021.

\bibitem{yang2022vision}
Jinyu Yang, Jiali Duan, Son Tran, Yi~Xu, Sampath Chanda, Liqun Chen, Belinda
  Zeng, Trishul Chilimbi, and Junzhou Huang.
\newblock Vision-language pre-training with triple contrastive learning.
\newblock In {\em The IEEE Conference on Computer Vision and Pattern
  Recognition (CVPR)}, 2022.

\bibitem{PointContrast2020}
Saining Xie, Jiatao Gu, Demi Guo, Charles~R Qi, Leonidas Guibas, and Or~Litany.
\newblock Pointcontrast: Unsupervised pre-training for 3d point cloud
  understanding.
\newblock In {\em Computer Vision--ECCV 2020: 16th European Conference,
  Glasgow, UK, August 23--28, 2020, Proceedings, Part III 16}, pages 574--591.
  Springer, 2020.

\bibitem{rozenberszki2022language}
David Rozenberszki, Or~Litany, and Angela Dai.
\newblock Language-grounded indoor 3d semantic segmentation in the wild.
\newblock In {\em Proceedings of the European Conference on Computer Vision
  ({ECCV})}, 2022.

\bibitem{qi_2019_votenet}
Charles~R Qi, Or~Litany, Kaiming He, and Leonidas~J Guibas.
\newblock {Deep Hough} voting for {3D} object detection in point clouds.
\newblock In {\em {Proceedings of the IEEE International Conference on Computer
  Vision}}, 2019.

\bibitem{Vaswani2017}
Ashish Vaswani, Noam Shazeer, Niki Parmar, Jakob Uszkoreit, Llion Jones,
  Aidan~N Gomez, {\L}ukasz Kaiser, and Illia Polosukhin.
\newblock {Attention} is all you need.
\newblock In {\em Advances in Neural Information Processing Systems}, pages
  5998--6008, 2017.

\bibitem{dosovitskiy2020vit}
Alexey Dosovitskiy, Lucas Beyer, Alexander Kolesnikov, Dirk Weissenborn,
  Xiaohua Zhai, Thomas Unterthiner, Mostafa Dehghani, Matthias Minderer, Georg
  Heigold, Sylvain Gelly, Jakob Uszkoreit, and Neil Houlsby.
\newblock {An} image is worth 16x16 words: {Transformers} for image recognition
  at scale.
\newblock {\em ICLR}, 2021.

\bibitem{oord2019representation}
Aaron van~den Oord, Yazhe Li, and Oriol Vinyals.
\newblock Representation learning with contrastive predictive coding.
\newblock {\em arXiv preprint arXiv:1807.03748}, 2019.

\bibitem{zhou_2018_open3d}
Qian-Yi Zhou, Jaesik Park, and Vladlen Koltun.
\newblock {Open3D}: {A} modern library for {3D} data processing.
\newblock {\em arXiv:1801.09847}, 2018.

\bibitem{kingma2014adam}
Diederik~P Kingma and Jimmy Ba.
\newblock Adam: A method for stochastic optimization.
\newblock {\em arXiv preprint arXiv:1412.6980}, 2014.

\bibitem{3detr}
Ishan Misra, Rohit Girdhar, and Armand Joulin.
\newblock {An} end-to-end {Transformer} model for {3D} object detection.
\newblock In {\em {ICCV}}, 2021.

\bibitem{bleu}
Kishore Papineni, Salim Roukos, Todd Ward, and Wei-Jing Zhu.
\newblock {B}leu: a method for automatic evaluation of machine translation.
\newblock In {\em Proceedings of the 40th Annual Meeting of the Association for
  Computational Linguistics}, pages 311--318, July 2002.

\bibitem{rouge}
Chin-Yew Lin.
\newblock {ROUGE}: A package for automatic evaluation of summaries.
\newblock In {\em Text Summarization Branches Out}, pages 74--81, Barcelona,
  Spain, July 2004.

\bibitem{meteor}
Alon Lavie and Abhaya Agarwal.
\newblock Meteor: An automatic metric for mt evaluation with high levels of
  correlation with human judgments.
\newblock In {\em WMT@ACL}, 2007.

\bibitem{cider}
Ramakrishna Vedantam, C.~Lawrence Zitnick, and Devi Parikh.
\newblock Cider: Consensus-based image description evaluation.
\newblock In {\em Proceedings of the IEEE Conference on Computer Vision and
  Pattern Recognition (CVPR)}, 2015.

\bibitem{mcan}
Zhou Yu, Jun Yu, Yuhao Cui, Dacheng Tao, and Qi~Tian.
\newblock {Deep} modular co-attention networks for visual question answering.
\newblock In {\em Proceedings of the IEEE Conference on Computer Vision and
  Pattern Recognition (CVPR)}, pages 6281--6290, 2019.

\bibitem{maaten_2008_t-sne}
Laurens van~der Maaten and Geoffrey Hinton.
\newblock {Viualizing data using t-SNE}.
\newblock {\em Journal of Machine Learning Research}, 9:2579--2605, 11 2008.

\end{thebibliography}
}


\end{document}